\documentclass{article} 

\usepackage[accepted]{icml2023}
\usepackage{amsmath,amsfonts,bm}

\def\eqref#1{equation~\ref{#1}}

\def\1{\bm{1}}

\DeclareMathAlphabet{\mathsfit}{\encodingdefault}{\sfdefault}{m}{sl}
\SetMathAlphabet{\mathsfit}{bold}{\encodingdefault}{\sfdefault}{bx}{n}

\DeclareMathOperator*{\argmin}{arg\,min}

\usepackage{hyperref}
\usepackage{url}

\usepackage[
	textsize=tiny,
	textwidth=3.2cm,
	colorinlistoftodos,
	backgroundcolor=teal!30!white,
	linecolor=teal!50!white
	]{todonotes}

\usepackage[utf8]{inputenc}

\usepackage{import}
\usepackage{hyperref}
\usepackage{url}
\usepackage{booktabs}
\usepackage{wrapfig, lipsum}
\usepackage{graphicx}
\usepackage{caption}
\usepackage{subcaption}
\usepackage{algorithm} 
\usepackage{algpseudocode} 

\usepackage[utf8]{inputenc} 
\usepackage[T1]{fontenc}    
\usepackage{hyperref}       
\usepackage{url}            
\usepackage{booktabs}       
\usepackage{amsfonts}       
\usepackage{nicefrac}       
\usepackage{microtype}      
\usepackage{xcolor}         
\usepackage{amsmath}
\usepackage{amsthm}
\usepackage{float}
\newcommand{\Exp}{{\mathbb{E}}}

\newtheorem{theorem}{Theorem}
\newtheorem*{theorem*}{Theorem}

\hypersetup{
	final,
	colorlinks,
	linkcolor=ourred,
	citecolor=ourblue
}
\definecolor{ourblue}{rgb}{0.368,0.507,0.71}
\definecolor{ourgreen}{rgb}{0.56,0.692,0.195}
\definecolor{ourred}{rgb}{0.923,0.386,0.209}

\icmltitlerunning{Semi-Autoregressive Energy Flows: Exploring Likelihood-Free Training of Normalizing Flows}

\begin{document}

\twocolumn[
\icmltitle{Semi-Autoregressive Energy Flows: \\Exploring Likelihood-Free Training of Normalizing Flows}




\icmlsetsymbol{equal}{*}

\begin{icmlauthorlist}
\icmlauthor{Phillip Si}{cornell,cmu}
\icmlauthor{Zeyi Chen}{tsinghua}
\icmlauthor{Subham Sekhar Sahoo}{cornell}
\icmlauthor{Yair Schiff}{cornelltech}
\icmlauthor{Volodymyr Kuleshov}{cornell,cornelltech}
\end{icmlauthorlist}

\icmlaffiliation{cornell}{Computer and Information Science, Cornell University, Ithaca, NY, USA}
\icmlaffiliation{cmu}{Machine Learning Department, Carnegie-Mellon University, Pittsburgh, PA, USA}
\icmlaffiliation{tsinghua}{Tsinghua University, Beijing, China}
\icmlaffiliation{cornelltech}{Department of Computer Science, Cornell Tech, NYC, NY, USA}

\icmlcorrespondingauthor{Phillip Si}{ps789@cs.cornell.edu}
\icmlcorrespondingauthor{Volodymyr Kuleshov}{kuleshov@cornell.edu}

\icmlkeywords{Machine Learning, ICML}

\vskip 0.3in
]
\printAffiliationsAndNotice{}


\begin{abstract}
Training normalizing flow generative models can be challenging due to the need to calculate computationally expensive determinants of Jacobians. 
This paper studies the likelihood-free training of flows and proposes the energy objective, 
an alternative sample-based loss based on proper scoring rules. 
The energy objective is determinant-free and supports flexible model architectures that are not easily compatible with maximum likelihood training, including semi-autoregressive energy flows, a novel model family that interpolates between fully autoregressive and non-autoregressive models. 
Energy flows feature competitive sample quality, posterior inference, and generation speed relative to likelihood-based flows; this performance is decorrelated from the quality of log-likelihood estimates, which are generally very poor.
Our findings question the use of maximum likelihood as an objective or a metric and contribute to a scientific study of its role in generative modeling.

\end{abstract}


\section{Introduction}

Normalizing flows 
form one of the major families of probabilistic and generative models \citep{rezende2016variational, kingma2016improved, papamakarios2019normalizing}.
They feature tractable inference and maximum likelihood learning and have applications in areas, such as image generation \citep{kingma2018glow}, anomaly detection \citep{nalisnick2019hybrid}, and density estimation \citep{papamakarios2017masked}.
However, training flows requires
calculating computationally expensive determinants of Jacobians; this constrains the range of architectures that can be used to parameterize flow models.

This paper questions the use of maximum likelihood for training normalizing flows and 
shows the existence of alternative objectives that do not require computing determinants or performing adversarial training \citep{grover2018flowgan}. These objectives yield highly performant models; however, their performance is entirely decorrelated from competitive log-likelihood scores, which are very poor.
Our findings contribute to the scientific understanding of generative model objectives and further question the common practice of training and evaluating models via the likelihood, hinting at the viability of alternative methods. 

Specifically, we introduce the {\em energy objective}, a sample-based multidimensional extension of proper
scoring rules that does not require computing model densities \citep{GneitingRaftery07}. 
The energy objective extends the recent autoregressive quantile flow framework of \citet{si2021autoregressive} to flexible non-autoregressive flow architectures. 
We complement this objective with efficient estimators based on random projections
and a theoretical analysis that draws connections to divergence minimization, and that highlights benefits over maximum likelihood. 


The energy objective enables training model architectures that are more flexible than ones trained using maximum likelihood (e.g., densely connected networks).
These models feature exact posterior inference (which is useful in applications such as semi-supervised learning and latent space manipulation), as well as exact likelihood estimation (which helps us study log-likelihood as an objective and a metric).
In particular, we propose semi-autoregressive flows, an architecture 
that interpolates between fully autoregressive and
non-autoregressive models.
From a practical perspective, energy flows achieve
improved sample quality, posterior inference,
and generation speed across a number of tasks and compared to equivalent model architectures trained using maximum likelihood.
Table \ref{tab:models} compares our approach to existing methods.


\paragraph{Contributions.}
In summary, this work (1) presents new results that question the use of maximum likelihood for training flows and proposes an alternative approach based on proper scoring rules and two-sample tests that extends quantile flows \citep{si2021autoregressive} to multiple dimensions. We (2) introduce specific two-sample objectives, such as the energy loss, and derive efficient slice-based estimators.
We also (3) provide a theoretical analysis for the proposed objectives as they are consistent estimators and feature unbiased gradients.
Finally, we (4) introduce a semi-autoregressive architecture with high speed and sample quality on generation and posterior inference tasks\footnote{Code is available at \href{https://github.com/ps789/SAEF}{https://github.com/ps789/SAEF}.}.

 \begin{table*}[t]
  \caption{Energy Flows and Semi-Autoregressive Energy Flows (SAEFs) are invertible generative models that feature expressive architectures, exact likelihood and posterior evaluation, and their training does not require computing log-determinants, in contrast to VAEs \citep{kingma2014stochastic}, MAFs  \citep{papamakarios2017masked}, NAFs \citep{huang2018neural}, AQFs \citep{si2021autoregressive}, GMMNets \citep{li2015generative}, and CramerGANs \citep{bellemare2017cramer}.}
  \label{tab:models}
  \centering
  \begin{tabular}{l|cccc}
    \hline
     Method & Likelihood / Posterior  & Sampling & Representation & Objective \\
    \hline
    VAE & Approx. & Feedforward & Gaussian & ELBO\\
    MAF & Exact & Autoregressive & Gaussian & Likelihood\\
    NAF & Exact & N/A & Non-Gaussian & Likelihood\\
    AQF  & Exact & Autoregressive & Non-Gaussian & Quantile Loss\\
    GMMNet & N/A & Feedforward & Non-Gaussian & MMD\\
    CramerGAN  & N/A & Feedforward & Non-Gaussian & Discr.+Energy\\
    \hline
    Energy Flow (Ours) & Exact & Feedforward & Non-Gaussian & Energy Loss\\
    SAEF (Ours) & Exact & Autoregressive & Non-Gaussian & Energy Loss\\
  \end{tabular}
 \end{table*}

\section{Background}



\paragraph{Normalizing Flow Models}


Generative modeling involves specifying a probabilistic model $p(\mathbf{y}) \in \Delta(\mathbb{R}^d)$ over a high-dimensional $\mathbf{y} \in \mathbb{R}^d$ \citep{kingma2014stochastic,goodfellow2014generative}.
%
A normalizing flow is a generative model $p(\mathbf{y})$ defined 
via an invertible mapping $f : \mathbb{R}^d \to \mathbb{R}^d$ 
between a noise variable $\mathbf{z} \in \mathbb{R}^d$ sampled from a prior $\mathbf{z} \sim p(\mathbf{z})$ and the target variable $\mathbf{y}$ \citep{rezende2016variational,papamakarios2019normalizing}. We may obtain an analytical expression for the likelihood $p(\mathbf{y})$ via the change of variables formula
$ p(\mathbf{y}) = \left| \frac{\partial f(\mathbf{z})^{-1}}{\partial \mathbf{z}} \right| p(\mathbf{z}),$
where $ \left| \frac{\partial f(\mathbf{z})^{-1}}{\partial \mathbf{z}} \right|$ denotes the determinant of the inverse Jacobian of $f$.
Computing this quantity is often expensive, hence we typically choose $f$ to be in a class of models for which the determinant is tractable \citep{rezende2016variational}, as in autoregressive models \citep{papamakarios2017masked}.


\paragraph{Proper Scoring Rules}

Consider a score or a loss $\ell : \Delta(\mathbb{R}^d) \times \mathbb{R}^d \to \mathbb{R}_+$
over a probabilistic forecast $F \in \Delta(\mathbb{R}^d)$ and a sample $\mathbf{y} \in \mathbb{R}^d$.
The loss $\ell$ is proper if the true distribution $G \in \argmin_F \mathbb{E}_{\mathbf{y} \sim G} \ell(F, \mathbf{y})$ \citep{GneitingRaftery07}. 
%
A popular proper loss is
the continuous ranked probability score (CRPS), defined for two cumulative distribution functions (CDFs) $F$ and $G$ as 
$
    \textrm{CRPS}(F, G) = \int \left(F(y) - G(y)\right)^2 dy.
$
When we only have samples from $G$, we can generalize this score to obtain the following loss for a single sample $y'$:
$
    \textrm{CRPS}_s(F, y') = \int_y \left(F(y) - \mathbb{I}(y-y')\right)^2 dy.
$
where $\mathbb{I}$ denotes the Heaviside step function.
The above CRPS can also be written as an expectation relative to the distribution $F$:
\begin{equation}\label{eqn:crps}
    \textrm{CRPS}(F, y') = -\frac{1}{2} \mathbb{E}_F |Y - Y'| + \mathbb{E}_F | Y - y'|,
\end{equation}
where $Y, Y'$ are independent copies of a random variable distributed according to $F$.
Recently, \citet{si2021autoregressive} proposed autoregressive quantile flows, which are trained using the CRPS and are determinant-free. We seek to extend the approach of \citet{si2021autoregressive} beyond autoregressive flows.

\paragraph{Two-Sample Tests and Integral Probability Metrics}

Two-sample tests compare distributions $F, G$ based on their respective sets of samples $\mathcal{D}_F = \{\mathbf{y}^{(i)}\}_{i=1}^m$ and $\mathcal{D}_G =\{\mathbf{x}^{(i)}\}_{i=1}^n$. Specifically, a two-sample test defines a statistic $T : \mathbb{R}^d \to \mathbb{R}$, and we determine whether $\mathcal{D}_F,\mathcal{D}_G$ originate from identical or different distributions $F,G$ based on differences in $T$ across $\mathcal{D}_F,\mathcal{D}_G$.
Two-sample tests motivate objectives for generative models such as generative moment matching networks (GMMNets; \citep{dziugaite2015training,li2015generative}) and generative adversarial networks (GANs; \citep{goodfellow2014generative}).
Two-sample tests are also an attractive training objective for flows because they are density-free and therefore do not require computing determinants \citep{grover2018flowgan}.

More modern approaches include integral probability metrics (IPMs) \citep{muller1997ipm}, which take the form $\max_{T \in \mathcal{T}} \mathbb{E}_{y \sim F} [T(y)] - \mathbb{E}_{y \sim G} [T(y)]$, where $\mathcal{T}$ is a family of functions.
A special case of IPMs is maximum mean discrepancy (MMD) \citep{gretton2008kernel}, in which $\mathcal{T} = \{T : ||T||_\mathcal{H} \leq 1\}$ is the set of functions with bounded norm in a reproducing kernel Hilbert space (RKHS) with norm $||\cdot||_\mathcal{H}$; the CRPS objective can be shown to be a form of MMD \citep{gretton2008kernel}.
\section{Exploring Determinant-Free Training of Normalizing Flows}



`We propose training normalizing flows using objectives inspired by two-sample tests, which do not require computing densities. This idea poses two sets of challenges: (1) most classical two-sample tests (e.g., Kolmogorov-Smirnov) are defined in one dimension and do not have simple multivariate extensions; (2) modern two-sample tests (e.g., IPMs) extend to high dimensions, but typically require solving a costly optimization problem.
Here, we derive two-sample tests that form good learning objectives, and we use the theory of proper scoring rules to justify their validity.


\subsection{Sample-Based Training of Normalizing Flows and the Energy Objective}


We seek to extend autoregressive quantile flows \citep{si2021autoregressive} to general architectures without the limitations of autoregressivity (e.g., slow sampling).
%
Specifically, we leverage a generalization of the sample-based form of the CRPS objective (\ref{eqn:crps}) to a multi-dimensional version called the {\em energy score} by \citet{szekely03,GneitingRaftery07}:
\begin{equation}\label{eqn:energy}
    \textrm{CRPS}_e(F, \mathbf{y}') = -\frac{1}{2} \mathbb{E}_F ||\mathbf{Y} - \mathbf{Y}'||_{\mathbf{2}}^\beta + \mathbb{E}_F || \mathbf{Y} - \mathbf{y}'||_{\mathbf{2}}^\beta,
\end{equation}
where $\beta \in (0,2)$, $||\cdot||_{\mathbf{2}}$ denotes the Euclidean norm, and $\mathbf{Y}, \mathbf{Y}' \in \mathbb{R}^d$ are independent copies of a vector-valued random variable distributed according to $F$.
%
The rightmost term $\mathbb{E}_F || \mathbf{Y} - \mathbf{y}'||_{\mathbf{2}}^\beta$ promotes samples $\mathbf{Y}$ from $F$ that are close to the data point $\mathbf{y'}$; the leftmost term $\frac{1}{2} \mathbb{E}_F ||\mathbf{Y} - \mathbf{Y}'||_{\mathbf{2}}^\beta$ encourages the model to produce diverse samples and not concentrate all probability mass on one $\mathbf{y}$.

\paragraph{The Kernelized Energy Objective}


As a generalized extension of the CRPS, the Kernelized Energy Objective extends the Euclidean norm to a kernel function:
\begin{equation}\label{eqn:kernelized_energy}
    \textrm{CRPS}_K(F, \mathbf{y}') = -\frac{1}{2} \mathbb{E}_F K(\mathbf{Y}, \mathbf{Y}') + \mathbb{E}_F K(\mathbf{Y}, \mathbf{y}'),
\end{equation}
The kernelized energy loss can be shown to be a proper loss \citep{GneitingRaftery07} and, thus, represents a valid training objective for a generative model. 
The flow objective consists of $\mathbb{E}_{\mathbf{y}' \sim \mathcal{D}} [\textrm{CRPS}_K(F, \mathbf{y}') ]$ and reveals the connection to two-sample tests between $F$ and $\mathcal{D}$.

\paragraph{Two-Sample Baselines}

In Appendix \ref{app:multidim_objectives}, we define two classical statistical tests as baselines and illustrate examples of alternative methods that can be derived from our two-sample-based approach.
In brief, {\bf Hotelling's two-sample test} uses the statistic $H_2(\mathcal{D}_F, \mathcal{D}_G) = (\mathbf{m}_F - \mathbf{m}_G)^\top S^{-1} (\mathbf{m}_F - \mathbf{m}_G)$, where 
$\mathbf{m}_F, \mathbf{m}_G$ are sample means, $S_F, S_G$  are sample variances and $S = (S_F + S_G)/2$. The {\bf Fr\'{e}chet distance} uses the objective $ R(\mathcal{D}_F, \mathcal{D}_G) = ||\mathbf{m}_F - \mathbf{m}_G||_2^2 
    + \mathrm{tr}(S_F + S_G - 2 (S_F S_G)^{1/2}),$ where we are using the same notation.

\subsection{Theoretical Properties}


\paragraph{Divergence Minimization}

When the variable $y \in \mathbb{R}$ is one-dimensional, the objective $\text{CRPS}(F,G)$ is precisely equivalent to the Cram\'{e}r divergence $\ell_2^2 (F, G) = \int_{-\infty}^\infty (F(y) - G(y)^2 dy$ between distributions $F, G$. 
\citet{szekely03} show that the one-dimensional version of the energy loss (\ref{eqn:energy}) is precisely equivalent to $\ell_2^2$. 
The kernelized version is a valid divergence between distributions, which directly follows from the fact that it is a proper loss \citep{GneitingRaftery07}. 
The connection to divergence minimization lends additional support to using (\ref{eqn:energy}) and (\ref{eqn:kernelized_energy}) as principled objectives---if $G$ is the data distribution, minimizing (\ref{eqn:energy}) or (\ref{eqn:kernelized_energy}) over a space of models produces an $F$ that is close to $G$.

\paragraph{Unbiased Gradient Estimation}

Our objectives have the property that given a sequence $\mathbf{Y}_n$ of $n$ samples $\mathbf{y}_1, \mathbf{y}_2, ..., \mathbf{y}_n$ from $G$, the gradient of the empirical distribution over these samples yields an unbiased estimate of the gradient of the expected loss: $\nabla_\theta \mathbb{E}_{\mathbf{Y}_n} \ell(F_\theta, \hat G_n) = \nabla_\theta \ell(F_\theta, G),$
where $\ell$ is one of our objective functions, $\hat G_n$ is the empirical distribution over 
$\mathbf{Y}_n,$ and $F_\theta$ is a model with parameters $\theta$ that we are optimizing. The statement above follows directly from the fact that both the energy and the CRPS objectives are proper scoring rules \citep{bellemare2017cramer}.


\paragraph{Why Energy Objectives?}

Consider the set of objectives $\ell_p^p (F, G) = \int_{-\infty}^\infty (F(y) - G(y))^p dy$ for $p \geq 1$ over $y \in \mathbb{R}$; these are also known as Wasserstein $p$-metrics \citep{kantorovich1939wasserstein}. The energy objective corresponds to $\ell_2^2$, and it is {\em the only} $\ell_p^p$ objective to support unbiased gradients \citep{bellemare2017cramer}.
In high dimensions, IPMs are general-purpose two-sample tests; popular IPMs include the  Kantorovich metric \citep{kantorovich}, Fortet-Mourier metric \citep{ASENS_1953_3_70_3_267_0}, the Lipschitz (or Dudley) metric \citep{dudley1966}, and the total variation distance.
In general, IPMs are defined in terms of a potentially costly optimization problem; out of the aforementioned IPMs, only the energy objective has a known analytical (optimization-free) solution, and it also features a faster statistical convergence rate \citep{sriperumbudur2009ipm}.

Overall, we summarize the above facts as part of the following formal result:
\begin{theorem}\label{thrm:energy}
The energy objectives (\ref{eqn:energy}) and (\ref{eqn:kernelized_energy}) are consistent estimators for the data distribution and feature unbiased gradients. 
\end{theorem}
This follows from properties of proper scoring rules and MMD; see Appendix \ref{app:proof} for the full proof.

\subsection{Scaling Sample-Based Flow Objectives Using Projections}

The framework of IPMs provides a wide range of high-dimensional sample-based objectives \citep{muller1997ipm}.
However, most of these objectives involve costly optimization problems, with the energy loss being a rare exception.
At the same time, there exist many popular one-dimensional two-sample tests that have appealing statistical and computational properties and can yield training objectives.

We propose further improving our objective via {\em random projections}, specifically {\em slicing}, which projects data into one dimension \citep{kolouri2019slicedwasserstein, song2019sliced, nguyen2020distributionalsliced}. Formally, we define a sampling probability $p(\mathbf{v})$ over one-dimensional vectors $\mathbf{v} \in \mathbb{R}^d$. We define a sliced version of a one-dimensional loss function $L_p(x, y) : \mathbb{R} \times \mathbb{R} \to \mathbb{R}$ as
\begin{equation}
    L_p(\mathbf{x}, \mathbf{y}) = \mathbb{E}_{\mathbf{v} \sim p(\mathbf{v})}
    \left[ 
    L(\mathbf{v}^\top \mathbf{x}, \mathbf{v}^\top \mathbf{y})
    \right].
\end{equation}
We approximate the expectation with Monte-Carlo samples.

\paragraph{Sliced Energy Objectives}

The sliced energy objective applies (\ref{eqn:crps}) to the projected data.
In practice, we find that the number of slices needed for good performance is lower than the dimensionality of the data, resulting in a favorable computational profile. Furthermore, we can formally prove that the resulting objective has appealing statistical properties.


\begin{theorem}
The sliced versions of the energy objectives (\ref{eqn:energy}) and (\ref{eqn:kernelized_energy}) are consistent estimators for the data distribution and feature unbiased gradients.
\end{theorem}

Intuitively, the first part of the theorem is true because the CRPS objective is related to the MMD.
At the same time, for each $\mathbf{v}$ the objective remains a proper score; a weighted combination of proper scores is also a proper score, hence the second part holds.
See Appendix \ref{app:proof} for the full proof.
Recall also that in one dimension, the energy loss reduces to the CRPS, which is equivalent to the Wasserstein-2 distance. Wasserstein distances have more favorable convergence properties \citep{arjovsky2017wgan} than maximum likelihood training, which lends further support to our choice of objective.

\paragraph{Sliced Two-Sample Baselines}

Slicing also allows us to use univariate two-sample tests as objectives. We describe several objectives in Appendix \ref{app:sliced_objectives}. In brief, these include: {\bf Kolmogorov-Smirnov}, a popular statistical test defined for two CDFs $F$ and $G$ as $\text{KS}(F, G) = \sup_{y} | F(y) - G(y) |$; {\bf Hotelling's $t^2$} test $H_u(\mathcal{D}_F, \mathcal{D}_G) = \frac{(m_F - m_G)^2}{s^2},$ a sliced version of Hotelling's objective; the sliced version of the {\bf Fr\'{e}chet} objective  $R_u(\mathcal{D}_F, \mathcal{D}_G) = (m_F - m_G)^2 + (s_F^2 - s_G^2)^2$.

\section{Energy Flows}

Next, we introduce {\em energy flows}, a class of models trained with our proposed determinant-free objectives.
An energy flow is defined by an invertible mapping between $\mathbf{z}$ and $\mathbf{y}$ and is trained using the energy loss.
As a result, energy flows improve over classical flow models by, among other things,
supporting flexible architectures and by simultaneously providing fast training and sampling.

Previous work on normalizing flows involved constrained architectures with tractable determinants, such as autoregressive models. In contrast, our model supports flexible feedforward architectures, which we outline in Appendix \ref{app:architectures}. In brief, our loss supports {\bf dense invertible flows} (DIFs), sequences of fully connected layers constrained to be invertible, {\bf invertible residual networks} \citep{behrmann2019invertible}, as well as {\bf rectangular flows} (REFs), in which the dimensionality of $\mathbf{y}$ and $\mathbf{z}$ is not equal \citep{nielsen2020survae,pmlr-v130-cunningham21a,caterini2021rectangular}.

\subsection{Motivation for Energy Flows}

Energy flows possess the following useful features: (1) exact posterior inference; (2) fast feed-forward generation; (3) a stable and effective training objective; (4) implicitly defined distributions (i.e., $p(x\mid z)$ is not assumed to be of any parametric form); (5) flexibility in terms of architecture for parameterizing the model. Closest in terms of features to energy flows are Generative Moment Matching Networks (GMMNets \citep{dziugaite2015training, li2015generative})---our work can be seen as introducing a principled way of doing posterior inference in GMMNets. Also related are VAEs; however they do not possess property (4), and in our experiments produce worse samples. Classical flows do not possess (5). GANs and autoregressive also differ in terms of training stability and generation speed, respectively.

It may appear that energy flows do not retain a key benefit of normalizing flows—being able to compare models via their log-likelihood. We argue that model comparison using log-likelihood may not be a good idea to being with; also, we see energy flows as a tool for the scientific study of the limitations of log-likelihood as a metric. 
Moreover, energy flows retain other aforementioned benefits over classical flows and other models: exact posterior inference, fast sampling via flexible architectures, and improved generation and latent space quality.

\citet{grover2018flowgan} introduced adversarial training of flows and showed that it yields poor log-likelihoods; we propose alternative objectives that are more stable while retaining competitive sample quality.
Our results contribute to the above line of work and further question the use of the likelihood for training flows. See Appendix \ref{motivation} for more details.

\subsection{Semi-Autoregressive Flows}


We also introduce semi-autoregressive flows (SAEF; pronounced ``safe''), an architecture trained with the energy loss that combines the speed
of feed-forward architectures with the sample quality of autoregressive models.
The SAEF model divides $d$-dimensional data into $B$ blocks and generates samples blockwise.
As a result, sampling time is reduced by a factor of $O(d/B)$ relative to autoregressive models.

Formally, SAEFs define an invertible mapping 
between a latent variable $\mathbf{z}$ and an observed variable $\mathbf{y}$ and 
require choosing a partition of $\mathbf{y}, \mathbf{z}$ into $B$ ordered blocks $(\mathbf{y}_b)_{b=1}^B$ and $(\mathbf{z}_b)_{b=1}^B$ (e.g., 4x4 blocks of pixels in an image).
They induce a probabilistic model $p(\mathbf{y})$ over $\mathbf{y}$ that factorizes as $p(\mathbf{y})=\prod_{b=1}^B p(\mathbf{y}_{b}|\mathbf{y}_{<b})$, where each $p(\mathbf{y}_{b}|\mathbf{y}_{<b})$ is defined via an invertible mapping
\begin{align}
\mathbf{y}_b = \tau(\mathbf{z}_b; \mathbf{h}_b) && \mathbf{h}_b = c_b(\mathbf{y}_{<b}),
\end{align}
where $\tau(\mathbf{z}_b; \mathbf{h}_b$) is an invertible transformer mapping the $b$-th latent block $\mathbf{z}_b$ to the $b$-th observed block $\mathbf{y}_b$, and $c_b$ is the $b$-th conditioner, which outputs transformer parameters $\mathbf{h}_b$.
Any invertible feed-forward energy flow can be used to parameterize $\tau$---we provide specific examples below.
The entire SAEF is trained via a sum of energy losses applied to each block 
$\mathbb{E}_{\mathbf{y} \sim \mathcal{D}}\left[ \sum_{b=1}^B \ell(F_{\mathbf{y}_{b}}, \mathbf{y}_b)\right],$
where $\mathcal{D}$ is a training set, $\mathbf{y} \sim \mathcal{D}$ is a datapoint sampled from $\mathcal{D}$, $F_{\mathbf{y}_{b}}$ is the distribution over $\mathbf{y}_{b}$ induced by $\tau(\cdot, \mathbf{h}_{b}(\mathbf{y}_{<b}))$, and $\ell$ is one of our two-sample losses, such as (\ref{eqn:energy}) or (\ref{eqn:kernelized_energy}).
    


When blocks are one-dimensional, this reduces to a standard autoregressive architecture that features high sampling quality but slow sampling speed. When blocks are full-dimensional, this reduces to a non-autoregressive energy flow with fast sampling but possibly worse quality.
In our experiments, we show that SAEFs can trade-off between these two regimes and obtain the best of both worlds.
Note also that SAEFs are hard to train using maximum likelihood, as they require specifying invertible non-autoregressive mappings $\tau$ between possibly high-dimensional blocks $\mathbf{y}_b, \mathbf{z}_b$---{\bf the SAEF architecture is only trainable using the energy objective}. See Appendix \ref{safpseudo} for pseudocode.

\section{Experiments}




We evaluate our framework on a range of UCI datasets \citep{Dua:2019}, datasets of handwritten digits \citep{scikit-learn,deng2012mnist}, and real world images, such as CIFAR10 \cite{krizhevsky2009learning} and Celeb-A \cite{liu2015faceattributes}.

\paragraph{Baselines}

We benchmark our models against normalizing flows trained using either maximum likelihood or quantile loss \cite{si2021autoregressive} and parameterized by either autoregressive or non-autoregressive architectures.
Our autoregressive models are based on baselines from earlier work \citep{papamakarios2017masked,si2021autoregressive} and include Autoregressive Quantile Flows (AQF) and Masked Autoregressive Flows (MAF-LL). These models assume a parameterization
$p(\mathbf{y} | \mathbf{z}) = \prod_{j=1}^d p(y_j | y_{<j}, z_j),$
where each $p(y_j | y_{<j})$ is a probability conditioned on the previous variables and $z_j$. In MAFs, the $p(y_j | y_{<j})$ are Gaussian; in AQFs they are parameterized by a flexible quantile flow \citep{si2021autoregressive}. 
We also compared to models trained with a Jacobian-free objective.
Specifically, we train autoregressive MAF and AQF models with the quantile loss \citep{si2021autoregressive} and denote these as MAF-QL and AQF-QL, respectively.

Our non-autoregressive baselines consist of variational auto-encoders (VAEs) trained using the evidence lower bound (ELBO) on the maximum likelihood and based on a fully-connected architecture (see below for details).
In order to understand the benefits of our objective, we fit a VAE model with the same invertible architecture for the decoder as the one used by our energy flow models; we refer to the resulting method as a dense invertible flow trained using maximum log-likelihood (DIF-LL).
We additionally compare against two state-of-the-art flow models, FFJORD \citep{grathwohl2018ffjord} and Invertible Resnets \citep{behrmann2019invertible}, using their respective open-source codebases. 

\paragraph{Energy Flow Models}
We constructed flow models using dense invertible layers, referring to the resulting model as a Dense Invertible Flow trained with an energy loss (DIF-E).
The DIF-E model consists of three feedforward invertible layers and Leaky ReLU activation functions and is trained using the kernelized energy loss (\ref{eqn:kernelized_energy}) with a mixture of RBF kernels with bandwidth in $\{2, 5, 10, 20, 40, 80\}$. 
We also use the energy score to train non-invertible rectangular flows trained with the energy loss (REF-E). In particular, rectangular flows are parametrized by layers of size $[d/8, d/4, d/2, d]$ as compared to DIF-E which requires all layers to be of size $d,$ for invertibility.

\paragraph{Metrics}

We evaluate the models in terms of log-likelihood (when available and appropriate) and using variants of the CRPS metric. For VAE-type models trained using the ELBO, we report the ELBO as a lower bound on the log-likelihood. We use two CRPS-style metrics which have the following structure: the first is a sample-based version as in (\ref{eqn:energy}).
The second, marked as univariate CRPS (U-CRPS), is the sum of one-dimensional CRPS measured for each output dimension and estimates the quality of marginal distributions \citep{si2021autoregressive}. 
Both versions use the $\ell_1$ norm.
See Appendix \ref{app:def_crps} for more details.
In our image datasets, we are also interested in a quantitative estimate of the quality of the generated samples and their similarity to the data distribution.
We therefore define a metric called the D-loss, which is measured by the accuracy of a discriminative model in determining whether an image is generated (see Appendix \ref{app:dloss} for details).

\subsection{Understanding Sample-Based Objectives}\label{subsec:sample_based}

We start with experiments that analyze the properties of the sliced energy objectives and compare them to baseline two-sample objectives on the UCI and image datasets.

\paragraph{Energy Objective vs. Two-Sample Baselines}

We claim that the energy objective is a particularly favorable training criterion for flows.
We empirically establish this fact by comparing it against the other two-sample objectives, which we see as strong baselines.
We train an invertible flow model on the Miniboone UCI dataset.
Complexity is written where $b$ denotes the batch size, $d$ denotes the dimension, and $n$ denotes the number of projections made.

\begin{table*}[t]
\centering
\caption{CRPS and Complexity for invertible flows trained on Miniboone UCI dataset with various two-sample objectives.
KS stands for Kolmogorov-Smirnov test and FD stands for Fr\'{e}chet Distance.}
\begin{tabular}{c||c|c|c|c|c|c|c}\hline
Metrics & KS &	1D Hotelling &	Hotelling &	1D-FD & FD & 1D-Energy & Energy\\ \hline
CRPS  &	1.53 &	0.57 & 0.717 &	0.558 &	0.559 & 0.545 &	0.548\\
Complexity & $n\log{b}$ &	$n$ & $d^3$ & 	$n$ & 	$d^3$ & $bn$ & $bd$\\
\end{tabular}
\label{tab:slicing_survey}
\end{table*}
Flows trained using the energy objective achieve the best performance in Table \ref{tab:slicing_survey}, outperforming the strong baselines.
In subsequent experiments, we focus on the energy objective.

\paragraph{Improvements in Scalability From Slicing}

\begin{table}[h]
\centering
\caption{Slicing on MNIST.}
\begin{tabular}{c||c|c|c|c}\hline
$n$  &	400  &	200	 & 100 &  50 \\
\hline
U-CRPS & 0.088	& 0.088	& 0.088 &	0.091\\
CRPS & 0.191 & 0.191 &	0.192 & 0.195\\
\hline
Block-$n$  &	100  &	20	 & 10 &  5 \\
\hline
U-CRPS & 0.084	& 0.085	& 0.084 &	0.084\\
CRPS & 0.086 & 0.087 &	0.086 & 0.087\\
\end{tabular}
\label{tab:slicing_projection}
\end{table}
Next, we seek to understand the scalability improvements from slicing.
We train a projection based model on MNIST using projected energy loss. The model consists of 4 dense layers of size 784 with Leaky ReLU activation functions for the first three layers and a sigmoid activation function for the last. When we calculate the loss, we take $n$ projections into a single dimension, which we denote in the top half of Table \ref{tab:slicing_projection} as $n$ for the projection parameters.
We additionally conduct slicing experiments on the SAEFs, in particular the 7x7 block size variant.
Slicing is conducted separately for each block in a similar manner to the fully-feedforward flow, so we are projecting a 49-dimensional block down to a single dimension.
The number of projections per block is denoted in Table \ref{tab:slicing_projection} as Block-$n$. In general, even for the block parameterization, CRPS is stable, even when using fewer projections.

We see in Table \ref{tab:slicing_projection} that sliced objectives perform comparably to non-sliced objectives while having improved computational complexity.

\paragraph{On Likelihood vs. Non-Likelihood Based Losses} 
Our work questions the use of maximum likelihood for training flow models.
In Table \ref{tab:glow_mnist}, we observe that a Glow model \cite{kingma2018glow} trained with likelihood has low CRPS, and conversely poor likelihoods when trained with an energy loss, though samples have high quality for both setups (see Appendix \ref{app:mnist_samples}, Figure \ref{fig:glow_samples} for sample generations).



\begin{table}[h]
\centering
\caption{Glow on MNIST.}
 \begin{tabular}{c||c|c}\hline
Models & NLL & CRPS\\
\hline
Glow-LL & 2.57 & 0.197\\
Glow-Energy & 7.03 & 0.190\\
\end{tabular}
\label{tab:glow_mnist}
\vspace{-10pt}
\end{table}

\subsection{UCI Experiments: Use of CRPS as a Metric}\label{subsec:uci}
In the previous sections, we primarily use CRPS as an evaluation metric for the performance of models trained with different losses.
In this section, we compare CRPS to other evaluation metrics.
Specifically, we implement energy flows on the Miniboone and Hepmass UCI datasets, which have been used previously as benchmarks by \citet{papamakarios2017masked} and \citet{si2021autoregressive} 
We use an LSTM architecture \cite{hochreiter1997long} for all autoregressive models.
In Table \ref{tab:uci_crps}, we report Frechet Distance, MMD, and CRPS and find that they are strongly correlated.
See Appendix \ref{expanded_uci} for CRPS-based evaluation on an expanded set of UCI datasets.

We note that training and evaluating with the same metric is not uncommon: in fact, most generative models are trained and evaluated using the log-likelihood.

\begin{table}
    \caption{Various Metrics on the UCI Datasets}
    \label{tab:uci_crps}
    \centering
    \begin{tabular}{llll}
    \hline
        \textit{Miniboone} & ~ & ~ & ~ \\ 
        Method & Frechet Distance & MMD & CRPS \\ \hline
        MAF-LL & 3.97 & 1.392 & 0.561 \\
        MAF-QL & 3.14 & 1.408 & 0.567 \\
        DIF-E & 1.62 & 0.088 & 0.524 \\ \hline
        \textit{Hepmass} & ~ & ~ & ~ \\
        Method & Frechet Distance & MMD & CRPS \\ \hline
        MAF-LL & 0.532 & 1.235 & 0.617 \\
        MAF-QL & 0.501 & 1.179 & 0.614 \\
        DIF-E & 0.274 & 0.072 & 0.58 \\
    \end{tabular}
\end{table}

\subsection{MNIST}

On MNIST, we train autoregressive models with the PixelCNN architecture. The architecture has 3 residual blocks, and each masked convolutional layer has a kernel size of 7.
Our SAEF models use the same general architecture, but adapted for the corresponding block sizes and trained using the block energy loss. The PixelCNN \cite{2016oordgated} maps from a base distribution (either a normal distribution, PixelMAF, or a uniform distribution, PixelAQF-QL) to the target distribution.
SAEF models utilize the same architecture but with generation sectioned off into different blocks.
In addition, we train a rectangular flow model using the ELBO approximation of the log-likelihood (REF-LL), resulting in a model equivalent to a VAE. Results are shown in Table \ref{tab:mnist_experiments}. Samples for a state-of-the-art flow model (FFJORD) and SAEF-4 are presented in Figure \ref{fig:mnist_samples} (see Appendix \ref{app:full_mnist} for more models).
\begin{figure}
\begin{subfigure}{.5\textwidth}
    \centering
    \includegraphics[width=7cm]{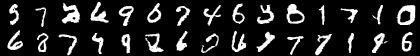}
    \caption{FFJORD (64)}
    \label{fig:mnist-ffjord}
\end{subfigure}
\begin{subfigure}{.5\textwidth}
    \centering
    \includegraphics[width=7cm]{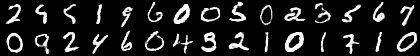}
    \caption{SAEF-4}
    \label{fig:mnist_saef}
\end{subfigure}
\vspace{-20pt}
\caption{MNIST samples}
\label{fig:mnist_samples}
\vspace{-15pt}
\end{figure}

\begin{table*}
  \caption{MNIST Generation Experiments}
  \centering
  \begin{tabular}{lrrrrrrr}
    \toprule
    Method  & U-CRPS  & CRPS & D-Loss & FID & MMD & Training {\footnotesize (sec)} & Sampling {\footnotesize (sec)}\\
    \midrule
    PixelMAF-LL & .128 & .279 &  1.00 & 100.55& 0.296 & 35 & 195.38\\
    PixelMAF-QL & .099 & .215 & .983 & 85.08 & 0.287 & 35 & 195.38\\
    PixelAQF-QL & .119 & .228 &  .986 & 61.27  & 0.232 & 29 & 195.38\\
    REF-LL (VAE) & .090 & .189 & .903 & 44.55 & 0.140 & 8 & 0.25\\
    DIF-LL (VAE) & .089 & .190 & .855 & 36.91 & 0.131 & 10 & 0.48\\
    FFJORD (16) & .101 & .208 & .650 & 24.78 & 0.103 & 540 & 48.88\\
    FFJORD (64) & .102 & .209 & .633 & 9.69 & 0.087 & 3100 & 155.69\\
    iResNet & .100 & .206 & .642 & 41.47 & 0.111 & 840 & 2.43\\
    Flow-GAN & .085 & .187 & 0.608 & 43.67 & 0.068 & 15 & 0.40\\
    GLOW & .090 & .197 & 0.983 & 63.06 & 0.600 & 1400 & 190.63\\
    \midrule
    REF-E & .085 & .187 & .778 &  41.04 & 0.052 & 3 & 0.21\\
    DIF-E & \textbf{.084} & \textbf{.186} & .701 & 22.76 & \textbf{0.051} & 6 & 0.40\\
    DIF-E-Proj & .085 & .186 & .819 & 22.55 & 0.056 & 3 & 0.40\\
    \midrule
    SAEF-2 & .085 & .188 &  .675 & 9.86 & 0.167 & 32 & 31.22\\
    SAEF-4 & .085 & .187 &  \textbf{.567} & \textbf{7.05} & 0.081 & 12 & 8.19\\
    SAEF-7 & .085 & .187 & .608 & 14.91 & 0.088 & 6 & 2.17\\
    SAEF-14 & .085 & .187 & .650 & 19.57 & 0.068 & 5 & 0.93\\
    \bottomrule
  \end{tabular}
  \label{tab:mnist_experiments}
 \end{table*}
\paragraph{Results} Though the fully feedforward (DIF-E) and sliced (DIF-E-Proj) variants demonstrate decent performance with fast training and generation speed and good CRPS, we find that introducing a degree of autoregressive modeling, improves results.
In particular, SAEF-4 greatly outperforms these non-autoregressive energy-based models in terms of FID.
In addition, 
SAEF-4 provides an order of magnitude sampling time speedup compared to some fully autoregressive models. 

\paragraph{Inversion and Interpolation.}
\begin{figure}
    \includegraphics[width = 8cm]{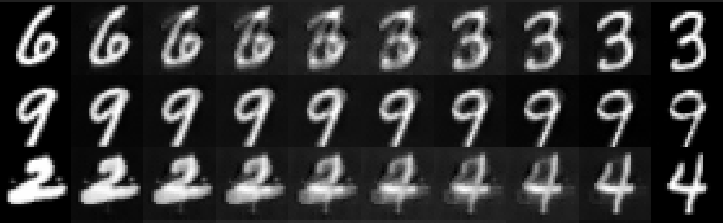}
    \caption{Interpolated MNIST samples with Energy Flow}
    \label{fig:interpolate}
\end{figure}

A key feature of the DIF-E model is exact posterior inference (despite not being trained with log-likelihood).
To demonstrate this, we 
create intermediate representations between pairs of MNIST samples through which we can smoothly interpolate (Figure \ref{fig:interpolate}).
Inverting the decoder model, by inverting the activation functions and weight matrices, allows us to create an encoder, similar to that of the VAE.
Like the VAE, the energy flow can generate interpolated samples, with the added advantage of exact posterior inference.

Second, we examine the utility of latent space of DIF-E by training a logistic regression model to predict the class of each datapoint based on its latent representation $\mathbf{z}$.
We split the original 10,000 test samples into 8,000 training data and 2,000 evaluation data for the logistic model.
Compared to a standard LL-trained Glow model's latent representations which produce 84.7\% accuracy, our energy-trained Glow model yields improved accuracy of 88.3\%. 

\subsection{CIFAR-10}
We modify a PixelCNN model \citep{2016oordgated} to (a) use the energy objective and (b) use a semi-autoregressive architecture.
In Table \ref{tab:cifar_experiments}, we see that Energy-based training of a standard PixelCNN yields equal or slightly better sample quality.
Further modifying the architecture yields improvements in sampling speeds at a cost of a relatively small reduction in image quality.
Samples are depicted in Figure \ref{fig:pixelcnn_samples_small} (with additional samples in Appendix \ref{app:cifar_samples}).

 \begin{table}
  \caption{CIFAR Generation Experiments}
  \label{tab:cifar_experiments}
  \centering
  \begin{tabular}{lll}
    \toprule
    Method  & FID & Sampling (sec) \\
    \midrule
    PixelCNN \citep{2016oordgated}  & 65.93 & 489\\
    \midrule
    PixelCNN-Energy-1x & 63.95 & 489\\
    PixelCNN-Energy-2x & 74.51 & 312\\
    PixelCNN-Energy-4x & 81.33 & 98\\
    
    \bottomrule
  \end{tabular}
 \end{table}

\begin{figure}[h]
\begin{subfigure}{.5\textwidth}
    \centering
    \includegraphics[width=7.5cm]{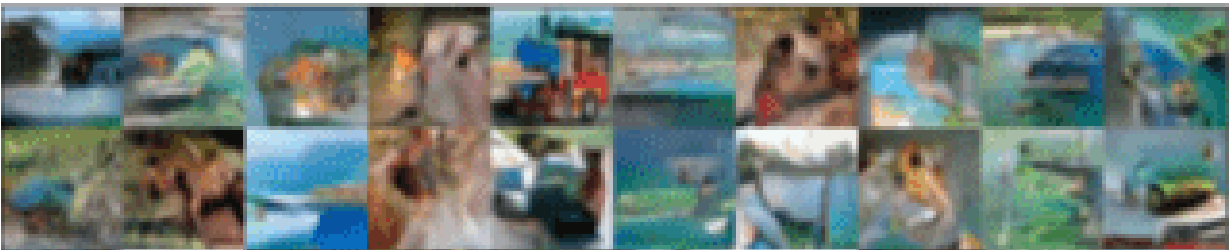}
    \caption{PixelCNN-Energy-1x}
    \label{fig:block_1_cifar}
\end{subfigure}
\begin{subfigure}{.5\textwidth}
    \centering
    \includegraphics[width=7.5cm]{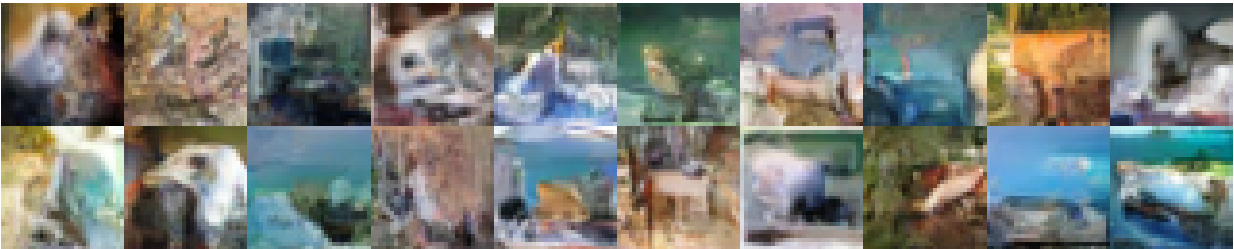}
    \caption{PixelCNN-Energy-2x}
    \label{fig:blended}
\end{subfigure}
\begin{subfigure}{.5\textwidth}
    \centering
    \includegraphics[width=7.5cm]{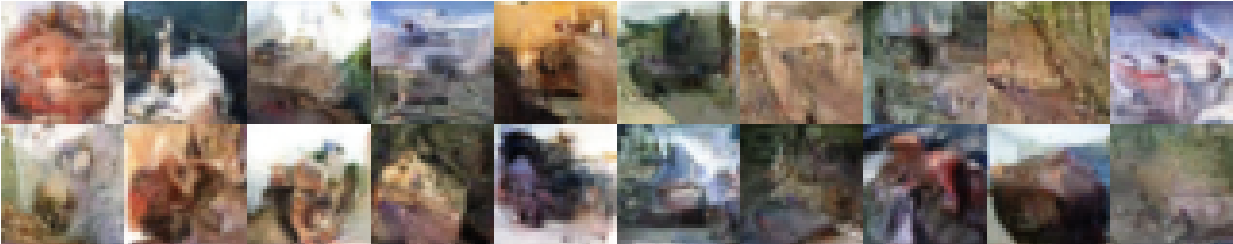}
    \caption{PixelCNN-Energy-4x}
    \label{fig:block_2_cifar}
\end{subfigure}
\begin{subfigure}{.5\textwidth}
    \centering
    \includegraphics[width=7.5cm]{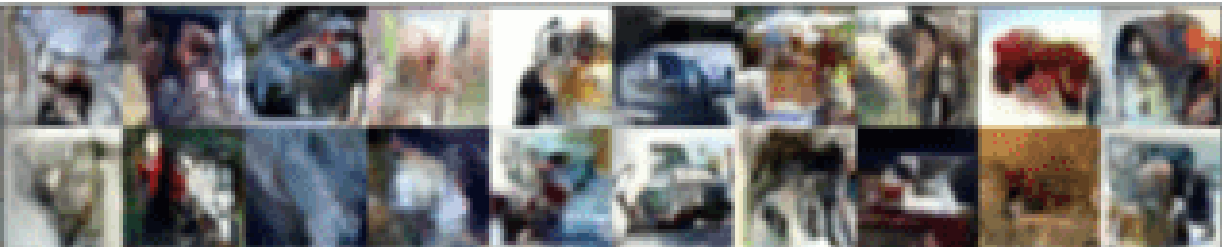}
    \caption{Gated PixelCNN}
    \label{fig:pixelcnn_ll}
\end{subfigure}
\caption{CIFAR samples}
\label{fig:pixelcnn_samples_small}
\end{figure}

\subsection{Celeb-A}
We also include an experiment on the Celeb-A dataset where we follow the same methodology as in our CIFAR-10 experiment.
We can see in Table \ref{tab:celeba_experiments} that switching to the energy loss for the PixelCNN considerably improves FID.
Even the 2x2 block version shows marked improvements in generation quality, with a much faster sampling time (75 seconds compared to 540 seconds for 10k samples) and improves over other baselines.
For comparison, we also include Glow \citep{kingma2018glow} and VAE \citep{kingma2013autoencoding} baselines trained with LL, with FID values obtained from \citet{xie2022tale}.

\begin{table}[h]
    \caption{Celeb-A Generation Experiments}
    \label{tab:celeba_experiments}
    \centering
    \begin{tabular}{lll}
    \hline
        Method & FID \\ \hline
        VAE  & 38.76 \\
        Glow & 23.32\\
        PixelCNN & 36.17\\ \hline
        PixelCNN-Energy-1x & 16.23\\
        PixelCNN-Energy-4x & 18.73\\ \hline
    \end{tabular}
    \vspace{-10pt}
\end{table}


\section{Discussion and Related Work}




\paragraph{Comparison to Other Generative Model Families}

We provide a complete comparison to other models in Table \ref{tab:models}. Our approach contrasts against VAE style models \citep{kingma2014stochastic} by providing exact inference and likelihood evaluation.
Unlike MAFs \citep{papamakarios2017masked}, our models are neither fully autoregressive in nature, nor do they make any Gaussianity assumptions.
They instead use a fully neural parameterization of the output probabilities, which contributes to improved performance and modeling flexibility.
Approaches like AQF \cite{si2021autoregressive} and NAF \cite{huang2018neural} also provide neural approximators but use fully autoregressive architectures, resulting in slow sampling.
Closely related feedforward models include GMMNets \citep{li2015generative,dziugaite2015training} and CramerGANs \citep{bellemare2017cramer}, but they do not offer likelihood evaluation and posterior inference.
Our framework is also related to the Maximum Mean Discrepancy (MMD) \citep{gretton2008kernel} and its generalizations (\cite{li2015generative}, \cite{dziugaite2015training}). 
However, unlike models optimizing MMD \citep{li2015generative,dziugaite2015training}, ours provide latent variable inference and exact density evaluation using the change of variables formula.

We also draw comparisons to self-normalizing flows \cite{keller2021self} and flow matching \cite{lipman2022flow}.
Self-normalizing flows \cite{keller2021self} achieve a similar goal to ours via two neural networks, parameterizing the forward and backward directions separately.
Unlike our work, their method yields an approximate inverse (albeit one that works well in practice) and relies on a training objective that could be more susceptible to local optima, e.g., if the two neural networks are imperfect inverses of each other.
Their method has the advantage of providing good scalability and modularity, as flow components can be composed to arbitrary depth and the training objective decomposes per step, while in our framework all components are trained together, which might not scale as well with depth.
Flow matching \cite{lipman2022flow} is a concurrent work based on continuous normalizing flows.
It can be seen as an extension of diffusion models to more general corruption processes.
Thus, it is a determinant-free approach to flows, similar to stochastic differential equation-based formulations of diffusion models.
It inherits the advantages of diffusion models, such as high-quality samples and good log-likelihoods; disadvantages include slow sampling speeds.

\paragraph{Comparison to Other Architectures}

Coupling layers are an invertible architecture that is an alternative to autoregressive flows.  A coupling layer partitions the input into two blocks and keeps one block the same while defining the output of the second block as an invertible transformation conditioned on the first block.
Our semi-autoregressive architecture sits between the coupling layers and autoregressive flows: it makes larger modifications to the input than a coupling layer, but smaller ones than an autoregressive layer.
Crucially, our semi-autoregressive layer has significantly fewer restrictions on its parametric form than a classical coupling layer: the latter typically implements a scale-and-shift operation (e.g., in Glow \cite{kingma2018glow}, RealNVP \cite{dinh2017density}), while our semi-autoregressive layer admits much more expressive transformations when it is trained with the energy loss.

\paragraph{The Likelihood as an Objective and Metric}

Our work explores the benefits and limitations of the likelihood as an objective and metric for generative models. 
We proposed energy flows, a model trained with an alternative sample-based objective, but whose likelihood is still tractable. We found that energy flows yield worse likelihoods while being faster and producing better samples (as measured by FID). This adds to an existing body of work on the decorrelation between likelihood and sample quality \citep{grover2018flowgan}, and extends it to non-adversarial objectives. 
We visualized samples from a number of energy flows; our qualitative examination suggests that mode collapse plays a role in poor likelihood scores at least in some settings (e.g., Figure \ref{fig:pixelcnn_samples_small}).
Thus, while our work provides a more stable likelihood-free objective than that of \citet{grover2018flowgan}, it also suggests that mode collapse is caused by the fact that an objective is sample-based, not the fact that it is adversarial. Other sample-based models may thus also suffer from mode collapse \citep{dziugaite2015training,li2015generative}.

Our findings add further evidence to the notion that if one cares about sample quality, the likelihood is not the ideal objective or metric; however there may be a price to pay for high quality samples, such as diversity. 
The right tradeoff between quality and diversity is a choice that needs to be made by the user; our work adds tools for making this choice.
More broadly, our work suggests the potential for further research into alternative training objectives for generative models; these can include regularization, even using likelihood as in \citet{grover2018flowgan}.

\section{Conclusion}
In this work, we proposed training normalizing flows using the energy objective, a proper scoring rule that does not require computing determinants during training. We then proposed semi-autoregressive energy flows, which feature fast sampling and high sample quality. Using an invertible flow architecture also allows us to retain exact posterior inference in energy flows.
We see our work as questioning the use of likelihood for training normalizing flows, and a a further step in exploring determinant-free flows based on novel learning objectives and architectures.

\section*{Acknowledgements}
This work was supported by Tata Consulting Services, the Hal \& Inge Marcus PhD Fellowship, and an NSF CAREER grant (\#2145577).
\bibliography{sources}
\newpage
\appendix
\onecolumn
\section{Additional Experimental Details} \label{sec:experimental}
\subsection{Architecture Details}
\paragraph{Abbreviations and Losses}
In Table \ref{tbl:abbrevs}, we provide a list of abbreviations, where the upper half consists of the abbreviation for the model names, and the bottom half consists of the abbreviations for losses.

Throughout the text we use loss abbreviations appended to model abbreviations to indicate which the objective used for was training a particular model (e.g., PixelMAF-LL corresponds to a PixelMAF model trained with log likelihood).
For the remaining baseline models which do not follow this convention (FFJORD, GLOW, and iResNet, in Table \ref{tab:mnist_experiments}) we note that they are trained with log-likelihood, and note that Flow-GAN uses an adversarial loss combined with a log likelihood term.

\begin{table}[ht!]
\begin{center}
\caption{Abbreviations for models (upper) and losses (lower) used throughout the text.}\label{tbl:abbrevs}
\begin{tabular}{l|c}
\toprule
Abbreviation & Full Name\\
\midrule
\multicolumn{2}{l}{\textit{Models}}  \\
AQF & Autoregressive Quantile Flow \\
DIF & Dense Invertible Flow \\
MAF & Masked Autoregressive Flow\\
REF & Rectangular Flow\\
SAEF & Semi-Autoregressive Energy Flow\\
iResNet & Invertible Residual Network\\
Pixel & PixelCNN\\
\midrule
\multicolumn{2}{l}{\textit{Losses}} \\
E & Energy Loss\\
LL & Log Likelihood\\
QL & Quantile Loss\\
\bottomrule
\end{tabular}
\end{center}
\end{table}

\paragraph{Slicing}
For our slicing experiments with the various multidimensional and single-dimension losses, the invertible feedforward model consists of 4 layers of size 43 (to match Miniboone dimension), and each objective is trained for 200 epochs with a learning rate of 1e-3. Each sample is projected onto 200 random normal vectors, and the resulting two-sample loss is summed across the 200 projections.
\paragraph{UCI Experiments}
For the MAF and AQF models, the LSTM architectures are composed of two LSTM layers with hidden size equal to the dimensionality of the data.
All models were trained with a batch size of 200 and learning rate of $1\mathrm{e}^{-3}$ using the ADAM optimizer \cite{kingma2014adam} for 200 epochs for the smaller datasets (Miniboone, Hepmass) and 20 epochs for the larger ones (Gas, Power, BSDS 300).
\paragraph{Image Generation Experiments}
For the PixelCNN architecture, we used a receptive field of 7 for digits and a receptive field of 15 for MNIST. We chose to switch the autoregressive architectures here from LSTM-based models used in the UCI experiments, which are more sequential, to the convolutional PixelCNN architecture to account for spatial location of features.
The PixelMAF-LL and PixelMAF-QL autoregressively transform pixels from the image into a distribution of the next pixel, except that PixelMAF-LL uses log likelihood loss while PixelMAF-QL uses quantile loss.
The PixelAQF-QL model takes in samples from a uniform distribution, and predicts the corresponding quantile loss for each pixel.

Our DIF models are parametrized with a layer size of 64 on the digits dataset, and 784 on MNIST, with invertible activation functions (Leaky ReLU activations up until the last layer and Sigmoid for the final activation).
The images are flattened prior to being passed through the feedforward model.
On MNIST, each model was trained for 300 epochs with a learning rate of $1\mathrm{e}^{-3},$ while on digits, each model was trained for 2,000 epochs.

The additional baselines are constructed as follows: each model is adapted from the code from their original papers.
FFJORD models use two blocks of stacked CNF layers composed of ODE Nets.
We tested with both a hidden size of 64 and a hidden size of 16, and trained for a total of 100 epochs with a learning rate of $1\mathrm{e}^{-3}.$
The Invertible ResNet \cite{behrmann2019invertible} has three scale blocks, each having 32 Invertible ResNet blocks, consisting of 32 filters with three convolution types with an ELU activation function \cite{clevert2015fast}.
It used an AdaGrad optimizer with a batch size of 128 trained for 70 epochs at a learning rate of $3\mathrm{e}^{-3}.$
Glow was trained with 3 levels and a depth of 1 on 8 GPUs for 250 epochs.



\subsection{D-Loss}\label{app:dloss}
The D-Loss is derived from an ensemble of SVM discriminators (with RBF kernels having a bandwidth parameter $\gamma$ of 0.0001, 0.001, 0.01, 0.1, 1, 10, 100, 1000, and 10000) used to differentiate the real data versus the sampled data from the model.
The D-Loss for a single discriminator model $D: \mathbb{R}^n \rightarrow \{0, 1\}$ is measured as the accuracy with which D can discern the difference between generated samples and true samples, which are labeled 0 and 1, respectively.
Given validation set with $m$ data points and labels $Y \in \{0, 1\}$, we have that
$$\text{D-Loss}_\gamma = \frac{1}{m}\sum_{i}^m \text{D}_\gamma(x^{(i)}) y^{(i)} + (1-\text{D}_\gamma(x^{(i)})) (1-y^{(i)})$$
The final D-Loss is measured by the validation accuracy achieved by the best SVM, which is gotten through a 80:20 train-validation split on the 300 real and 300 fake images:
$\text{D-Loss} = \max_{\gamma} \text{D-Loss}_\gamma$

\subsection{Defining CRPS and U-CRPS}\label{app:def_crps}
For simulated samples $\mathbf{x_1},...,\mathbf{x_m}$ and true data $\mathbf{y}$, we define CRPS as
\begin{align*}
\text{CRPS}((\mathbf{x_1},...,\mathbf{x_m}), \mathbf{y}) = 
\frac{1}{m}\sum_{i = 1}^m ||\mathbf{x_i} - \mathbf{y}||^2 - \frac{1}{2m^2} \sum_{i = 1}^ m \sum_{j = 1}^m ||\mathbf{x_i} - \mathbf{x_j}||^2    
\end{align*}

and U-CRPS as
\begin{align*}
\text{U-CRPS}((x_1,...,x_m), y) = \frac{1}{m}\sum_{i = 1}^m |\mathbf{x_i} - \mathbf{y}| - \frac{1}{2m^2} \sum_{i = 1}^ m \sum_{j = 1}^m |\mathbf{x_i} - \mathbf{x_j}| 
\end{align*}

\section{Theoretical Analysis and Proofs}\label{app:proof}

\paragraph{Multi-Variate Objectives}

In this section, we make the assumption that the kernels $K$ used to define our objectives are measurable and bounded by $\kappa$. Under these conditions, when using a kernelized objective with kernel $K$, each distribution $F$ can be represented by a mean embedding $\mu_F$ in the reproducing kernel Hilbert space (RKHS) induced by $K$ \citep{gretton2008kernel}. We also assume that there exists a unique mapping between $\mu_F$ and $F$ for every $F$ in the class of model distributions $\mathcal{F}$. Note that this can be satisfied for any Borel probability measure if the kernel $K$ is chosen to be universal or characteristic \citep{gretton2008kernel}. Alternatively, we may satisfy this claim by choosing our $\mathcal{F}$ to be a restricted set of distributions for which the above claim is true.

\begin{theorem*}
The energy objectives (\ref{eqn:energy}) and (\ref{eqn:kernelized_energy}) are consistent estimators for the data distribution and feature unbiased gradients.
\end{theorem*}

\begin{proof}

First, we seek to establish the consistency of the minimizer of our objectives as an estimator of the data distribution.
Our argument uses the fact that the (kernelized) energy score is closely connected to the maximum mean discrepancy (MMD; \citep{gretton2008kernel}). Observe that
\begin{align*}
    \mathbf{E}_{y' \sim \mathcal{D}} \textrm{CRPS}_K(F, \mathbf{y}') 
    & = \frac{1}{2} \mathbb{E}_F K(\mathbf{Y}, \mathbf{Y}') - \mathbb{E}_{F, \mathcal{D}} K(\mathbf{Y}, \mathbf{y}') \\
    & = ( \frac{1}{2} \mathbb{E}_F K(\mathbf{Y}, \mathbf{Y}') - \mathbb{E}_{F, \mathcal{D}} K(\mathbf{Y}, \mathbf{y}') - \frac{1}{2} \mathbb{E}_{y, y' \sim \mathcal{D}} K(\mathbf{y}, \mathbf{y}') ) + \frac{1}{2} \mathbb{E}_{y, y' \sim \mathcal{D}} K(\mathbf{y}, \mathbf{y}') \\
    & = - \frac{1}{2} \text{MMD}_K^2(F,\mathcal{D}) + \text{const}.
\end{align*}
Hence, by maximizing the CRPS over a set of possible models $F$, we are minimizing a monotonic transformation of the MMD. Since (\ref{eqn:energy}) is a special case of (\ref{eqn:kernelized_energy}), with the distance kernel \citep{sejdinovic2013equivalence}, the above claim also holds for the non-kernelized energy objective (\ref{eqn:energy}).

We would like to establish that minimizing objectives (\ref{eqn:energy}) and (\ref{eqn:kernelized_energy}) over a data distribution $\mathcal{D}_n$ of $n$ samples from the true data distribution $\mathcal{P}$ yields a model $F_n$ that is similar to what we would obtain if we searched for the best $F$ using the full data distribution $\mathcal{P}$; in other words:
$$
\mathbb{E}[L(F_n, \mathcal{P})]  \leq \inf_{F \in \mathcal{F}} L(F, \mathcal{P}) + o(n),
$$
where 
$L(F,\mathcal{P})$ 
is a metric or pseudo-metric\footnote{A metric $d(x,y)$ satisfies four properties: (1) symmetry, (2) the triangle inequality, (3) $d(x,x)=0,$ (4) and $d(x,y) =0 \Longleftrightarrow x = y$. A pseudo-metric satisfies only the first three properties. The MMD objective is a metric if its kernel is characteristic or universal (or more generally if there is a one-to-one mapping between $\mu_F$ and $F$); otherwise, it is a pseudo-metric.}
that we will instantiate shortly, $\mathcal{F}$ is the hypothesis class for the model $F$, $F_n$ is the empirical risk minimization solution (from our method) over a dataset $\mathcal{D}_n$, and the additive $o(n)$ term decays to zero as we increase $n$. 
Note that if the model is well-specified (i.e., $\mathcal{P} \in \mathcal{F}$), we have $\mathbb{E}[L(F_n, \mathcal{P})] = o(n)$,
and we have a consistent estimator.

To establish this fact, we will derive a version of the above identity for a modified version of the MMD, under the assumption of this section.
We will also argue that the kernelized energy estimate satisfies that identity.

First, let $L(F,\mathcal{P}) = \text{MMD}(F,\mathcal{P})$.
By the properties of MMD and kernels, we know that $\text{MMD}(F,\mathcal{P}) = ||\mu_F - \mu_\mathcal{P}||_\mathcal{H}$, and MMD is a pseudo-metric.
Note that we have by the triangle inequality
$$
L(F_n, \mathcal{P}) \leq L(F_n, \mathcal{D}_n) + L(\mathcal{D}_n, \mathcal{P}),
$$
where we overload notation and use $\mathcal{D}_n$ to also denote the empirical distribution.
Note that because our objective is a monotonic transformation $(\frac{1}{2}\text{MMD}^2+\text{const})$ of the MMD, the $F_n$ minimizes the MMD within $\mathcal{F}$. Thus we can write for any $F \in \mathcal{F}$
\begin{align*}
L(F_n, \mathcal{P}) 
& \leq L(F, \mathcal{D}_n) + L(\mathcal{D}_n, \mathcal{P}) \\
& \leq L(F, \mathcal{P}) + 2 L(\mathcal{D}_n, \mathcal{P})
\end{align*}
where we have used once more the triangle inequality in the last line. Taking expectations on both sides and using the fact that $F \in \mathcal{F}$ was arbitrary, we find that
\begin{align*}
\mathbb{E} L(F_n, \mathcal{P}) 
& \leq \inf_{F \in \mathcal{F}} L(F, \mathcal{P}) + 2 \mathbb{E} L(\mathcal{D}_n, \mathcal{P}) \\
& \leq \inf_{F \in \mathcal{F}} L(F, \mathcal{P}) + 2 \sqrt{\mathbb{E} L(\mathcal{D}_n, \mathcal{P})^2}.
\end{align*}
To establish our claim, we need to bound the last term. Let $x_i$ denote the i.i.d. samples from $\mathcal{D}_n$, let $\phi$ denote the embedding induced by the kernel $K$ in its RKHS $\mathcal{H}$, and note that we have
\begin{align*}
\mathbb{E} L(\mathcal{D}_n, \mathcal{P})^2
& = \mathbb{E} || \frac{1}{n} \sum_{i=1}^n \phi(x_i) - \mathbb{E} \phi(x) ||_\mathcal{H} \\
& = \text{Var}\Big(\frac{1}{n} \sum_{i=1}^n \phi(x_i)\Big) \\
& = \frac{1}{n} \text{Var}(\phi(x_1)) \\
& \leq \frac{2}{n} \mathbb{E} ||\phi(x_1)||_\mathcal{H} \\
& \leq \frac{2\kappa}{n}
\end{align*}
Thus, our main claim follows with
\begin{align*}
\mathbb{E} L(F_n, \mathcal{P}) 
\leq \inf_{F \in \mathcal{F}} L(F, \mathcal{P}) + 2 \sqrt{\frac{2\kappa}{n}}.
\end{align*}
Thus the estimated model $F_n$ satisfies the above inequality and if the data distribution $\mathcal{P} \in \mathcal{F}$, our consistency claim holds.

We can establish that the gradients  are unbiased by leveraging properties of proper scoring rules. Recall from the background section that a loss $L : \Delta(\mathbb{R}^d) \times \mathbb{R}^d \to \mathbb{R}$ is strictly proper \citep{GneitingRaftery07} if 
$G = \argmin_F \mathbb{E}_{\mathbf{y} \sim G} L(F, \mathbf{y})$. In the context of the CRPS objective $L$, we have by definition of a proper loss
$$
L(F,G) = \mathbb{E}_{\mathbf{y}' \sim G} \textrm{CRPS}_K(F, \mathbf{y}') 
= \mathbb{E}_{\mathbf{y}' \sim G} \textrm{CRPS}_K(F, G_{\mathbf{y}'}),
$$
where $G_{\mathbf{y}'}$ is the empirical distribution derived from $\mathbf{y}'$. 

Let $\mathcal{P}$ denote the true data distribution, $\mathcal{D}_n$ a dataset of size $n$ drawn from  $\mathcal{P}$, and $G_n$ the resulting empirical distribution.
Then we have:
\begin{align*}
    \nabla_\theta \mathbb{E}_{\mathcal{D}_n \sim \mathcal{P}} L(F_\theta, G_n))
    & =
     \nabla_\theta \mathbb{E}_{\mathcal{D}_n \sim \mathcal{P}} \mathbb{E}_{\mathbf{y}' \sim \mathcal{D}_n} L(F_\theta, \mathbf{y}') \\
    & =
     \nabla_\theta \mathbb{E}_{\mathbf{y}' \sim \mathcal{P}} L(F_\theta, \mathbf{y}') \\
    & =
     \nabla_\theta  L(F_\theta, \mathcal{P}),
\end{align*}
which is equivalent to the statement that we wanted to prove.

\end{proof}

\paragraph{Alternative Approaches to Showing Consistency}

The fact that consistency holds also follows from properties of the MMD for general classes $\mathcal{F}$; for example as shown in Dzuigaite et al. \citep{dziugaite2015training} (Theorem 1),
$$
\mathbb{E}[\text{MMD}^2(F_n, \mathcal{P})]  \leq \inf_{F \in \mathcal{F}} \text{MMD}^2(F, \mathcal{P}) + o(n),
$$
if $\mathcal{F}$ satisfies a fat-shattering condition. The desired consistency claim with $L(F,G)$ being our kernelized energy objective $\text{CRPS}_K$ then follows directly from our earlier derivation by applying an affine transformation on each side of the above equation. 

\paragraph{Alternative Approaches to Showing that Gradients are Unbiased}

Note that a special case of the unbiased gradient property for the non-kernalized objective (\ref{eqn:crps}) has been established using techniques discussed in \citet{bellemare2017cramer} (Proposition 3). This result also follows from our aforementioned connection to the MMD and Lemma 6 \citet{gretton2008kernel}.

\paragraph{Sliced Objectives}

\begin{theorem*}
The sliced versions of the energy objectives (\ref{eqn:energy}) and (\ref{eqn:kernelized_energy}) are consistent estimators for the data distribution and feature unbiased gradients.
\end{theorem*}

\begin{proof}
We establish the first part of the claim by observing that the sliced version of the energy objective


\begin{equation}
    \textrm{CRPS}(F, \mathbf{y}') = \mathbb{E}_{w \sim p(w)} \left[\frac{1}{2} \mathbb{E}_F K(w^\top \mathbf{Y}, w^\top \mathbf{Y}') - \mathbb{E}_F K(w^\top \mathbf{Y}, w^\top \mathbf{y}') \right],
\end{equation}
where $K$ is a kernel in 1D, is an affine transformation of squared MMD. Then the first part of the claim follows by the argument in Theorem \ref{thrm:energy}.
To see this, first, define the function $K_w : \mathbb{R}^d \times \mathbb{R}^d \to \mathbb{R}$ as
$$
K_w(x, y) = K(w^\top x, w^\top y),
$$
where $K$ is the kernel used as part of the sliced energy objective. It is easy to see that $K_w(x, y)$ is a kernel. Consider any dataset $S = \{x_i\}_{i=1}^k$; then the matrix $M$ defined as $M_{ij} = K_w(x_i, x_j)$ will be semi-definite because the corresponding matrix $M'$ defined as $M'_{ij} = K(w^\top x_i, w^\top x_j)$ is also positive definite, because it is the kernel matrix for the set $S = \{w^\top x_i\}_{i=1}^k$. Hence, by Mercer's theorem $K_w$ is a kernel.

Next, define the function
$\bar K : \mathbb{R}^d \times \mathbb{R}^d \to \mathbb{R}$ as
$$
\bar K(x, y) = \mathbb{E}_{w \sim p(w)} K_w(x, y).
$$
This is also a kernel, because it is a sum of kernels. Next, note that
\begin{align*}
    \textrm{CRPS}(F, \mathbf{y}')
    & = \mathbb{E}_{w \sim p(w))} \left[ \frac{1}{2} \mathbb{E}_F K(w^\top \mathbf{Y}, w^\top \mathbf{Y}') - \mathbb{E}_F K(w^\top \mathbf{Y}, w^\top \mathbf{y}') \right] \\
    & = \frac{1}{2} \mathbb{E}_F \bar K(\mathbf{Y}, \mathbf{Y}') - \mathbb{E}_{F, \mathcal{D}}
    \bar K(\mathbf{Y}, \mathbf{y}'),
\end{align*}
which is an instance of the kernelized energy objective that uses a modified kernel. Note that this is both a proper score and a rescaled version of the squared MMD with a modified kernel.
The two claims of this theorem follow directly from Theorem \ref{thrm:energy}.

\end{proof}

\section{Expanded Background on Proper Scoring Rules}


\subsection{Predictive Uncertainty in Machine Learning}

Probabilistic machine learning models predict a probability distribution over the target variable---e.g., class membership probabilities or the parameters of an exponential family distribution. 
We seek to produce models with accurate probabilistic outputs that are useful for generation.

\paragraph{Notation.}

Supervised models predict a target $y \in \mathcal{Y}$ from an input $x \in \mathcal{X}$.
, where $x, y$ are realizations of random variables $X, Y \sim \mathbb{P}$, and $\mathbb P$ is the data distribution.
We are given a model $H : \mathcal{X} \to \Delta_\mathcal{Y}$, which outputs a probability distribution $F(y) : \mathcal{Y} \to [0,1]$ within the set $\Delta_\mathcal{Y}$ of distributions over $\mathcal{Y}$; the probability density function of $F$ is $f$.
We are also given a training set $\mathcal{D} = \{ (x_i, y_i) \in \mathcal{X} \times \mathcal{Y} \}_{i=1}^n$
consisting of i.i.d.~realizations of  random variables $X, Y \sim \mathbb{P}$.

\subsection{Proper Scoring Rules}

Comparing point estimates from supervised learning models is straightforward: we can rely on metrics such as accuracy or mean squared error.
Probabilities, on the other hand, are more complex and require specialized metrics.

In statistics, the standard tool for evaluating the quality of predictive forecasts is a proper scoring rule \citep{gneiting2007probabilistic}. 
This paper advocates for evaluating the quality of uncertainties using proper scoring rules \citep{gneiting2007strictly}. 

Formally, let $L : \Delta_\mathcal{Y} \times \mathcal{Y} \to \mathbb{R}$ denote a loss between a probabilistic forecast $F \in \Delta_\mathcal{Y}$ and a realized outcome 
$y \in \mathcal{Y}$. Given a distribution $G \in \Delta_\mathcal{Y}$ over $y$, we use $L(F,G)$ to denote the expected loss
$
L(F,G) = \mathbb{E}_{y \sim G} L(F, y).
$

We say that $L$ is a {\em proper loss} if it is minimized by $G$ when $G$ is the true distribution for $y$:
$
L(F,G) \geq L(G,G) \text{ for all $F$}.
$
One example is the log-likelihood $L(F,y) = - \log f(y)$, where $f$ is the probability density or probability mass function of $F$. 
Another example is the check score for $\tau \in [0,1]$:
\begin{equation}\label{eqn:checkscore}
\rho_\tau(F, y) = 
\begin{cases}
\tau (y-F^{-1}(\tau)) & \text{ if $y \geq f$} \\
(1-\tau)(F^{-1}(\tau)-y) & \text{ otherwise}.
\end{cases}
\end{equation}
See Table \ref{tbl:properlosses} for additional examples.

What are the qualities of a good probabilistic prediction, as measured by a proper scoring rule?
It can be shown that 
every proper loss decomposes into a sum of the following terms \citep{gneiting2007probabilistic}:
$$\text{proper loss} = \text{calibration} \underbrace{- \text{sharpness} + \text{irreducible term}}_\text{refinement term}.$$
Thus, there are precisely two qualities that define an ideal forecast: calibration and sharpness.

\begin{table*}
\caption{Examples of three proper losses: the log-loss, the continuous ranked probability score (CRPS), and the quantile loss. A proper loss $L(F,G)$ between distributions $F, G$---assumed here to be cumulative distribution functions (CDFs)---decomposes into a calibration loss term $L_c(F,Q)$ (also known as reliability) plus a refinement term $L_r(Q)$ (which itself decomposes into a sharpness and an uncertainty term). Here, $Q(y)$ denotes the CDF of $\mathbb{P}(Y=y \mid F_X = F)$, and $q(y), f(y)$ are the probability density functions of $Q$ and $F$, respectively.}\label{tbl:properlosses}
\begin{center}
\begin{tabular}{l|c|c|c}
\toprule
{\bf Proper Loss } & {\bf Loss} & {\bf Calibration} & {\bf Refinement} \\
& $L(F,G)$ & $L_c(F,Q)$ & $L_r(Q)$ \\
\midrule
Logarithmic & $\Exp_{y\sim G}$ $\log f(y)$ & $\text{KL}(q||f)$ & $H(q)$ \\
CRPS & \hspace{2mm} {$\Exp_{y\sim G}$ $(F(y) - G(y))^2$} \hspace{2mm}& {$\int^{\infty}_{-\infty}(F(y) - Q(y))^2$dy} & \hspace{2mm} {$\int^{\infty}_{-\infty} Q(y) (1 - Q(y))dy$} \\
Quantile & {$\Exp^{\tau\in U[0,1]}_{y\sim G} \rho_\tau(F, y)$} & \hspace{2mm} {$\int_0^1 \int^{F^{-1}(\tau)}_{Q^{-1}(\tau)}(Q(y) - \tau)dy d\tau$} \hspace{2mm} & {$\Exp^{\tau\in U[0,1]}_{y\sim Q} \rho_\tau(Q, y)$} \\
\bottomrule
\end{tabular}
\end{center}
\end{table*}

\section{Details on Additional Two-Sample Training Objectives}
\label{app:objectives}

\subsection{Gaussian Two-Sample Baseline Objective Over High-Dimensional Vectors}
\label{app:multidim_objectives}

We use the following classical objectives as baselines for our work and to illustrate examples of alternative methods that can be derived from our two-sample-based approach; both tests make a Gaussian modeling assumption. 

\paragraph{Hotelling's Two-Sample Test}

Being
closely related to Student's t-test, it uses  the following statistic:
\begin{align}
    H_2(\mathcal{D}_F, \mathcal{D}_G) = (\mathbf{m}_F - \mathbf{m}_G)^\top S^{-1} (\mathbf{m}_F - \mathbf{m}_G),
\end{align}
where 
$\mathbf{m}_F = \frac{1}{m}\sum_{\mathbf{y}^{(i)} \in \mathcal{D}_F} \mathbf{y}^{(i)}$, 
$\mathbf{m}_G = \frac{1}{m}\sum_{\mathbf{y}^{(i)} \in \mathcal{D}_G} \mathbf{y}^{(i)}$ 
are the sample means, and the matrices
$S_F = \frac{1}{m-1}\sum_{\mathbf{y}^{(i)} \in \mathcal{D}_F} (\mathbf{y}^{(i)} - \mathbf{m}_F) (\mathbf{y}^{(i)} - \mathbf{m})^T$, 
$S_G = \frac{1}{m-1}\sum_{\mathbf{y}^{(i)} \in \mathcal{D}_G} (\mathbf{y}^{(i)} - \mathbf{m}_G) (\mathbf{y}^{(i)} - \mathbf{m})^T$
are sample covariances, while
$S = (S_F + S_G)/2$ is their average. This objective encourages the two samples to have similar means.

\paragraph{Fr\'{e}chet Distance}

This is another Gaussian-based distance that we use as an objective:
\begin{align}
    R(\mathcal{D}_F, \mathcal{D}_G) =
    &||\mathbf{m}_F - \mathbf{m}_G||_2^2 \nonumber \\
    &+ \mathrm{tr}(S_F + S_G - 2 (S_F S_G)^{1/2}),
\end{align}
where we are using the same notation as above. This objective is encouraging the model to produce data with similar means and variances. It is derived from the Fr\'{e}chet distance between two Gaussians.

\subsection{Sliced Two-Sample Baselines}
\label{app:sliced_objectives}

Slicing also allows us to use univariate two-sample tests as objectives. We give examples below.

\paragraph{Kolmogorov-Smirnov} One of the most popular ways of comparing the similarity between two distributions is via the quantity
\begin{equation}
    \text{KS}(F, G) = \sup_{y} | F(y) - G(y) |,
\end{equation}
the maximum distance between two CDFs $F$ and $G$.
Empirical CDFs can be used for samples. While this test corresponds to an IPM, it does not have a widely accepted extension to higher dimensions.

\paragraph{Hotelling's Univariate Objective}

The sliced version of Hotelling's objective corresponds to using Hotelling's $t^2$ univariate test (which is just the squared version of Student's $t$-test) as an objective.
\begin{align}
    H_u(\mathcal{D}_F, \mathcal{D}_G) = \frac{(m_F - m_G)^2}{s^2},
\end{align}
where $m_F, m_D$, and $s^2$ are respectively the sample mean of $\mathcal{D}_F$, the sample mean of $\mathcal{D}_G$, and the combined sample variance, defined as in the multivariate version.
Note that this formula is much less computationally expensive than the multi-dimensional one, which requires performing a matrix inversion (in worst-case $O(d^3)$ time), while the sliced version takes only $O(d)$ time.

\paragraph{Fr\'{e}chet Univariate Objective}

Similarly, the sliced version of the Fr\'{e}chet objective is written as:
\begin{align}
    R_u(\mathcal{D}_F, \mathcal{D}_G) = (m_F - m_G)^2 + (s_F^2 - s_G^2)^2,
\end{align}
which encourages the sample means $m_F, m_G$ and the sample variances $s_F, s_G$ to be the same.
Again, this $O(d)$ formula is less computationally expensive than in higher-dimensions, where it requires performing multiple matrix multiplications and a matrix square root ($O(d^3)$).

\section{Feed-Forward Architectures for Energy Flows.}\label{app:architectures}

Since our objective does not require computing Jacobians, we are able to use flexible classes of invertible models that are difficult to train using log-likelihood.
Previous work on invertible mapping leveraged integration-based transformers \citep{wehenkel2021unconstrained}, spline approximations \citep{muller2019neural, durkan2019neural, dolatabadi2020invertible}, piece-wise separable models, and others.
Our main requirement on the model architecture is efficient sampling.
We describe several feed-forward flow architectures that are compatible with our objective below.

\paragraph{Dense Invertible Layers}
The simplest architecture we consider consists of a sequence of small $\mathbb{R}^d \to \mathbb{R}^d$ dense layers with invertible non-linearities (such as $\tanh$ or Leaky ReLUs). We enforce the invertibility of the dense layers by adding a scaled identity component $\sigma I_d$ for small $\sigma > 0$; other options for inducing invertibility include positivity constraints on the weights \citep{huang2018neural}.
Although the $\tanh$ non-linearities are invertible, numerical values close to $\{-1, 1\}$ tend to introduce numerical instability during inversion.
We address this issue via activity regularization \citep{chollet2015keras}.
With these two architectural choices, for modestly sized $d$'s, we were able to compute both $\mathbf{z} \to \mathbf{y}$ and $\mathbf{y} \to \mathbf{z}$ mappings analytically and in a numerically stable way.

\paragraph{Invertible Residual Networks}

Recently, residual networks with spectral normalization have been proposed as a flexible invertible architecture \citep{behrmann2019invertible}. Although one of the two directions of the flow is not computable analytically, it may be approximated using a fixed-point iteration algorithm. Invertible residual networks are typically trained using maximum likelihood; computing the determinant of the Jacobian of each layer requires a sophisticated approximation based on Taylor series expansion. 
Interestingly, when training these flows using maximum likelihood, the "fast" direction needs to be $\mathbf{y} \to \mathbf{z}$ in order to enable fast training, but generation becomes non-analytic. In contrast, when training using an energy objective, the $\mathbf{z} \to \mathbf{y}$ direction is fast both for training and generation.

\paragraph{Rectangular Flow Architectures}

Recently, several authors explored rectangular flows, in which the dimensionality of $\mathbf{y}$ and $\mathbf{z}$ is not equal \citep{nielsen2020survae,pmlr-v130-cunningham21a,caterini2021rectangular}. Training with maximum likelihood involves sophisticated extensions to the change of variables formula.
However, these models can be trained without modification using an energy loss as long as one can sample from them efficiently.
At the same time, we may retain their pseudo-invertibility to perform posterior inference. 

\section{Pseudocode for Semi-Autoregressive Flows} \label{safpseudo}
In this section, we provide the algorithms for training (Algorithm \ref{algo:train}) and sampling from (Algorithm \ref{algo:sample}) SAEFs.

\begin{algorithm}[ht!]
	\caption{Semi-Autoregressive Energy Flow Training} 
	\begin{algorithmic}[1]
 \State \textbf{Input}: $\mathbf{x} \sim \mathcal{D}$
        \For {iteration $1,2,\ldots$}
		    \State Generate $\mathbf{z}$ from random normal $(0, 1)^d$.
			\For{$b=1,2,\ldots,Blocks$}
			    \State $\mathbf{h}_b = c_b(\mathbf{x}_{<b})$
				\State $\mathbf{y}_b = \tau(\mathbf{z}_b; \mathbf{h}_b)$
				\State $\mathbf{y'}_b = \tau(\mathbf{z'}_b; \mathbf{h}_b)$
				\State Compute Energy Loss($\mathbf{y}_b, \mathbf{y'}_b, \mathbf{x}_b$)\textbf{}
			\EndFor
			\State Backpropogate sum of energy loss on $\tau$
		\EndFor
	\end{algorithmic} 
        \label{algo:train}
\end{algorithm}

\begin{algorithm}[ht!]
	\caption{Semi-Autoregressive Energy Flow Sampling} 
	\begin{algorithmic}[1]
		\For {$b=1,2,\ldots,Blocks$}
		    \State $\mathbf{h}_b = c_b(\mathbf{y}_{<b})$
			\State $\mathbf{y}_b = \tau(\mathbf{z}_b; \mathbf{h}_b)$
		\EndFor\\
        Return $(\mathbf{y}_1,...,\mathbf{y}_b)$
	\end{algorithmic} 
 \label{algo:sample}
\end{algorithm}

\section{Additional Motivation for Energy Flows} \label{motivation}

From our experiments in Table \ref{tab:glow_mnist}, flows trained with the energy loss feature poor log-likelihoods.
However, the FID and CRPS metrics in Table \ref{tab:mnist_experiments} are good: these models still generate very good images.
We note that previous work has already shown that models trained with non-likelihood objectives tend to have very weak log-likelihoods,
e.g., training normalizing flows with adversarial losses \citep{grover2018flowgan}.
Our work adds additional results to this line of work.

We believe that exact posterior inference is a particularly important feature of flows. Popular papers (RealNVP \cite{dinh2017density}, Glow \cite{kingma2018glow}) use inferred latent $\mathbf{z}$ for interpolation, image manipulation, etc., which we have also done in our interpolation experiments.
Though these are not low-dimensional, high-dimensional $\mathbf{z}$ are still useful.
Many generative modeling families (GANs \cite{goodfellow2014generative}, PixelCNN/WaveNet \cite{oord2016wavenet}, GMMNets \cite{li2015generative}) do not have latent inference, and retaining this feature is an important advantage over many types of models.

\section{Glow Samples on MNIST}\label{app:mnist_samples}

\begin{figure}
\begin{subfigure}{.5\textwidth}
    \centering
    \includegraphics[width=7cm]{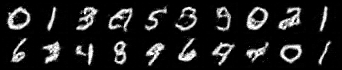}
    \caption{Energy}
    \label{fig:glow_energy}
\end{subfigure}
\begin{subfigure}{.5\textwidth}
    \centering
    \includegraphics[width=7cm]{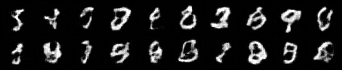}
    \caption{LL}
    \label{fig:glow_ll}
\end{subfigure}
\caption{Glow samples when trained with Energy and LL loss}
\label{fig:glow_samples}
\end{figure}

Samples for a Glow model trained with either the energy or a log-likelihood objective on the MNIST dataset are presented in Figure \ref{fig:glow_samples}.
As discussed, in Section \ref{subsec:sample_based}, relatively poor log-likelihood values do not translate to meaningful differences in generated sample quality.

\section{Autoregressive Samples on CIFAR10}\label{app:cifar_samples}
\begin{figure}
\begin{subfigure}{.5\textwidth}
    \centering
    \includegraphics[width=5cm]{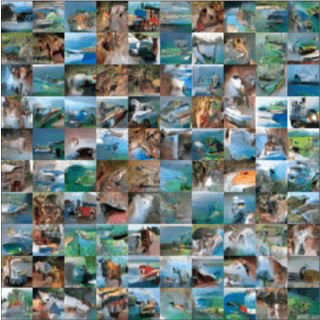}
    \caption{PixelCNN-Energy-1x}
    \label{fig:app_block_1_cifar}
\end{subfigure}
\begin{subfigure}{.5\textwidth}
    \centering
    \includegraphics[width=5cm]{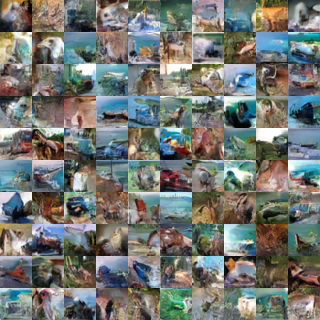}
    \caption{PixelCNN-Energy-2x}
    \label{fig:app_blended}
\end{subfigure}
\begin{subfigure}{.5\textwidth}
    \centering
    \includegraphics[width=5cm]{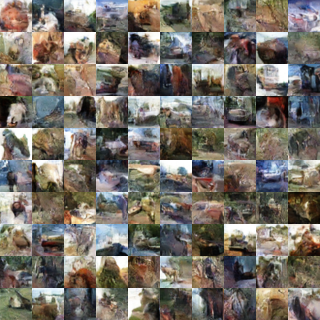}
    \caption{PixelCNN-Energy-4x}
    \label{fig:app_block_2_cifar}
\end{subfigure}
\begin{subfigure}{.5\textwidth}
    \centering
    \includegraphics[width=5cm]{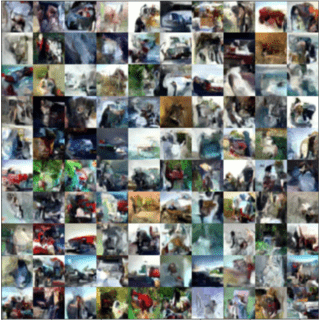}
    \caption{Gated PixelCNN}
    \label{fig:app_pixelcnn_ll}
\end{subfigure}
\caption{CIFAR samples}
\label{fig:pixelcnn_samples}
\end{figure}

Additional autoregressive samples for CIFAR10 are added in Figure \ref{fig:pixelcnn_samples}.
The blended (PixelCNN-Energy-2x) samples feature the first half of the image generated by the fully autoregressive SAEF-1, and the second half of the image generated by SAEF-2, which allows for a 30\% reduction in time compared to a fully-autoregressive model.

\section{Stability Comparison of MMD and GAN loss.}\label{app:stability}

\begin{figure}
    \centering
    \includegraphics[width=0.5\textwidth]{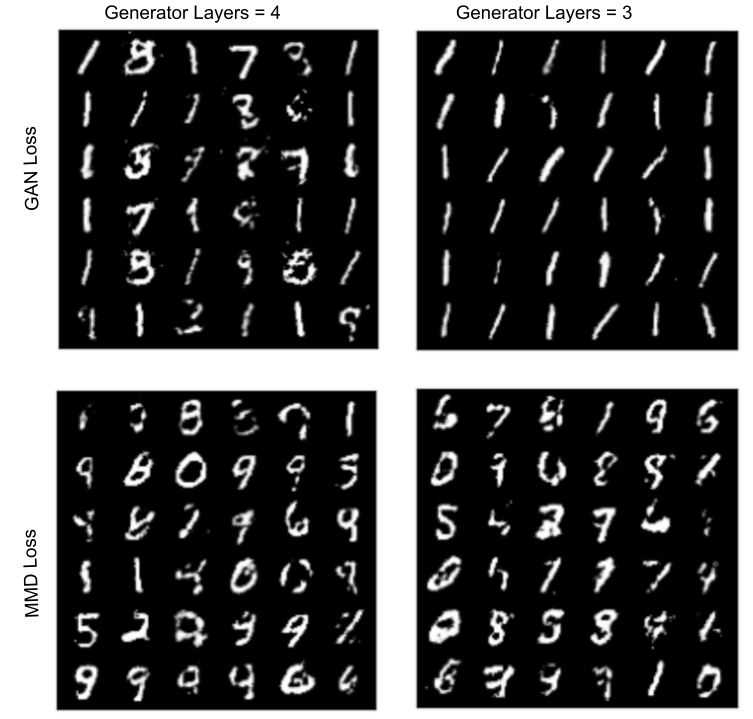}
    \caption{Stability comparision of the MMD and the GAN loss. A Generator network  trained using MMD loss typically requires the same hyperparameters to train the model when subtle changes are made to the network's architecture. On the right column we see that the sample quality deteriorated for a generator that was trained using the GAN loss after removing layer and used the optimal hyperaparameter configuration for a 4 layered network.}
    \label{fig:gan_vs_mmd}
\end{figure}

We study how robust a neural network architecture is to the changes in hyperparameter when it is trained using a adversarial loss vs MMD loss. The adversarial loss is defined using a convolutional discriminator network which tries to determine whether the data is a true sample or comes from the generator.
First, optimal hyperparameter settings for a generator network that features 4 fully connected layers of width 784 and Leaky ReLU activation were found: $lr = 2\mathrm{e}^{-5}$ for GAN loss and $lr=0.0002$ for MMD.
In Figure \ref{fig:gan_vs_mmd}, we observe that removing 1 extra layer from the generator causes a significant deterioration in the sample quality for a model trained using the GAN loss.
In contrast, when using MMD loss we do not see this deterioration.
For both settings, we train using the optimal hyperparameters for the corresponding loss functions.

\section{Additional MNIST Samples}\label{app:full_mnist}
\begin{figure}
\begin{subfigure}{.5\textwidth}
    \centering
    \includegraphics[width=7cm]{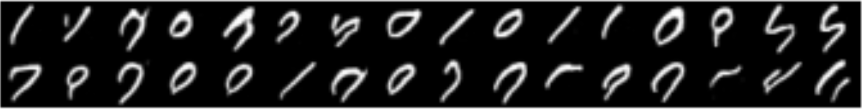}
    \caption{PixelMAF-LL}
    \label{fig:mnist_ll}
\end{subfigure}
\begin{subfigure}{.5\textwidth}
    \centering
    \includegraphics[width=7cm]{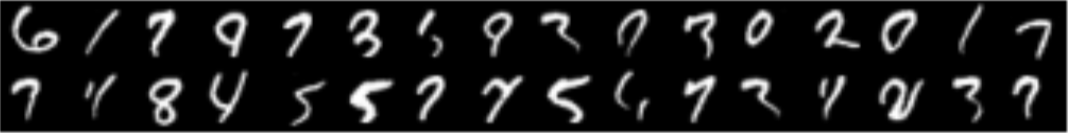}
    \caption{PixelMAF-QL}
    \label{fig:mnist_ql}
\end{subfigure}
\begin{subfigure}{.5\textwidth}
    \centering
    \includegraphics[width=7cm]{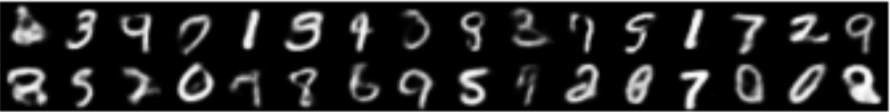}
    \caption{DIF-LL}
    \label{fig:mnist_vae}
\end{subfigure}
\begin{subfigure}{.5\textwidth}
    \centering
    \includegraphics[width=7cm]{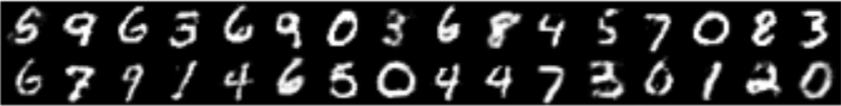}
    \caption{DIF-E}
    \label{fig:mnist_energy}
\end{subfigure}
\begin{subfigure}{.5\textwidth}
    \centering
    \includegraphics[width=7cm]{images/ffjord64.png}
    \caption{FFJORD (64)}
    \label{fig:app_mnist-ffjord}
\end{subfigure}
\begin{subfigure}{.5\textwidth}
    \centering
    \includegraphics[width=7cm]{images/block4.png}
    \caption{SAEF-4}
    \label{fig:app_mnist_saef}
\end{subfigure}
\vspace{-20pt}
\caption{MNIST samples from six methods}
\label{fig:full_mnist_samples}
\end{figure}
MNIST generated samples for a broader set of comparison models are provided in Figure \ref{fig:full_mnist_samples}.




\section{Expanded Results on UCI Evaluated by CRPS}\label{expanded_uci}
In Section \ref{subsec:uci} we presented results for two UCI datasets.
Here, we include CRPS evaluation on an expanded set of UCI datasets: BSDS 300, Miniboone, Gas, Power, and Hepmass, which have been used previously as benchmarks by \citet{papamakarios2017masked} and \citet{si2021autoregressive}.
The size of the datasets are noted in Table \ref{tab:uci_generative}.
The SAEFs use an appropriate block size $b$ (listed as a column in the table) which divides the dimension of the data. Miniboone bas been padded to ensure an even divisibility of the dimension of the dataset by the dimensionality of the block. This would apply to any other dimensionality for which the appropriate block size would not divide the dimensionality of the data.

\begin{table*}[h]
\centering
\caption{Model performance on UCI datasets as measured by CRPS.}
\begin{tabular}{c||c||c|c|c||c|c || c | c}\hline
\multicolumn{1}{c||}{Dataset} & $d$ & MAF-LL & MAF-QL & AQF-QL & DIF-E & DIF-E Proj & b & SAEF\\ \hline
BSDS 300  & 63 & .044 & .036 &	\textbf{.033} & .039 & .040 & 3 & .037\\
Miniboone  & 43 & .567 & .561	& .525 & .524 & .545 & 2 & \textbf{.521}\\
Gas  & 8 & .645 &	.565 &	\textbf{.513} & .548 & .551 & 2 & .530\\
Power  & 6 & .542 &	.506 &	.502 & .451 & .454 & 2 & \textbf{0.443}\\
Hepmass  & 21 & .617 &	.614 &	\textbf{.523} & .589 & .587 & 3 & .559\\
\hline
\end{tabular}
\label{tab:uci_generative}
\end{table*}

\paragraph{Results.} 
As shown in Table \ref{tab:uci_generative}, energy flows perform comparably to the neural AQF-QL baseline. Both methods are trained using variants of the CRPS and obtain top performance across the five datasets.
However, the DIF-E model is non-autoregressive, hence provides advantage in terms of sampling speed. Our experiment illustrate that non-autoregressive models can match the performance of autoregressive models trained with log-likelihood or versions of the CRPS objective.
SAEFs, which use a variant of the same loss as DIF-E, further improve upon its results, giving slightly better CRPS scores across the board, while still being reasonable in terms of sampling speed.

\section{Image Generation on Digits}
\begin{table*}
  \caption{Digits Generation Experiments}
  \label{tab:digits}
  \centering
  \begin{tabular}{lrrrrr}
    \toprule
    \cmidrule(r){1-2}
    Method     & U-CRPS  & CRPS & D-Loss & Training {\footnotesize (sec)} & Sampling {\footnotesize (sec)}\\
    \midrule
    PixelMAF-LL & 0.136 & 0.206 & 0.974 & 0.15 & 14.00 \\
    PixelMAF-QL & 0.131 & 0.204 & 0.883 & 0.15 & 14.00\\
    PixelAQF-QL & 0.127 & 0.199 & \textbf{0.681} & 0.13 & 14.00\\
    DIF-LL (VAE) & 0.138 & 0.207 & 0.941 & 0.07& 0.01\\
    \midrule
    REF-E & 0.127 & 0.201 & 0.823 & 0.12& 0.06\\
    DIF-E & \textbf{0.126} & \textbf{0.197}  & 0.807 & 0.14 & 0.07\\
    DIF-E-Proj & 0.127 & 0.199 & 0.815 & 0.06 & 0.08\\
    \midrule
    SAEF-1 & 0.126 & 0.198 & 0.795& 0.12 & 10.54\\
    SAEF-2 & 0.126& 0.199 & 0.754 & 0.07 & 2.73\\
    SAEF-4 & 0.127& 0.198 & 0.772 & 0.05 & 0.86\\
    \bottomrule
  \end{tabular}
\vspace{-5pt}
\end{table*}
We also test our methods on a small-scale generative modeling task: digit generation \citep{scikit-learn}, displaying results in Table \ref{tab:digits}.
We use a PixelCNN architecture for the autoregressive models, which we denote PixelMAF-LL, PixelMAF-QL, and PixelAQF-QL. The PixelCNN maps from a $(\mathbb{N}(0, 1))^d$ distribution (PixelMAF) and a $(\mathbb{U}(0, 1))^{d}$ distribution (PixelAQF-QL) to the target distribution. SAEF models also utilize the same PixelCNN architecture, but with its generation sectioned off into different blocks, each of which is evaluated by our energy loss.
We provide the ELBO loss as an upper bound on the NLL for VAE-type models, and training speed is given by seconds per epoch, while sampling speed is given by seconds per 1000 samples.

\paragraph{Results.}

The proposed DIF-E and SAEF models perform comparably on the U-CRPS and CRPS metrics to the AQF model, although the latter is more discriminable (has a better D-loss). On the other hand the samples generated by the DIF-E and SAEF outperforms those of any of the other autoregressive architectures, as well as the samples from non-autoregressive DIF-LL model, which is trained with maximum log likelihood.
\end{document}